\documentclass[sn-basic]{sn-jnl}

\usepackage{graphicx}%
\usepackage{multirow}%
\usepackage{multicol}%
\usepackage{amsmath,amssymb,amsfonts}%
\usepackage{amsthm}%
\usepackage{mathrsfs}%
\usepackage[title]{appendix}%
\usepackage{xcolor}%
\usepackage{textcomp}%
\usepackage{manyfoot}%
\usepackage{booktabs}%
\usepackage{algorithm}%
\usepackage{algorithmicx}%
\usepackage{algpseudocode}%
\usepackage{listings}%
\usepackage{natbib}
\usepackage{subcaption}
\usepackage{makecell}

\theoremstyle{thmstyleone}%
\theoremstyle{thmstyletwo}%

\theoremstyle{thmstylethree}%

\begin{document}

\title[Article Title]{Efficient Relational Context Perception for Knowledge Graph Completion}


\author[1,2,3]{\fnm{Wenkai} \sur{Tu}}\email{wenkaitu@whu.edu.cn}
\author[1,2,3]{\fnm{Guojia} \sur{Wan}}\email{guojiawan@whu.edu.cn}
\author[1]{\fnm{Zhengchun} \sur{Shang}}\email{zcshang@whu.edu.cn}
\author[1,2,3]{\fnm{Bo} \sur{Du}}\email{dubo@whu.edu.cn}

\affil[1]{\orgdiv{School of Computer Science}, \orgname{Wuhan University}, \orgaddress{\street{Luojiashan Road}, \city{Wuhan}, \postcode{430072}, \state{Hubei Province}, \country{China}}}

\affil[2]{\orgdiv{National Engineering Research Center for Multimedia Software, School of Computer Science}, \orgname{Wuhan University}, \orgaddress{\street{Luojiashan Road}, \city{Wuhan}, \postcode{430072}, \state{Hubei Province}, \country{China}}}

\affil[3]{\orgdiv{Institute of Artificial Intelligence, School of Computer Science}, \orgname{Wuhan University}, \orgaddress{\street{Luojiashan Road}, \city{Wuhan}, \postcode{430072}, \state{Hubei Province}, \country{China}}}

\abstract{Knowledge Graphs (KGs) provide a structured representation of knowledge but often suffer from challenges of incompleteness. To address this, link prediction or knowledge graph completion (KGC) aims to infer missing new facts based on existing facts in KGs.
Previous knowledge graph embedding models are limited in their ability to capture expressive features, especially when compared to deeper, multi-layer models. These approaches also assign a single static embedding to each entity and relation, disregarding the fact that entities and relations can exhibit different behaviors in varying graph contexts.
Due to complex context over a fact triple of a KG, existing methods have to leverage complex non-linear context encoder, like transformer, to project entity and relation into low dimensional representations, resulting in high computation cost.
To overcome these limitations, we propose Triple Receptance Perception (TRP) architecture to model sequential information, enabling the learning of dynamic context of entities and relations. Then we use tensor decomposition to calculate triple scores, providing robust relational decoding capabilities. This integration allows for more expressive representations.
Experiments on benchmark datasets such as YAGO3-10, UMLS, FB15k, and FB13 in link prediction and triple classification tasks demonstrate that our method performs better than several state-of-the-art models, proving the effectiveness of the integration. 
}




\maketitle

\section{Introduction}\label{sec1}

Knowledge Graphs (KG) \cite{nickel2015review} as an information system for providing well-structured real-world knowledge , are widely used in various applications such as question answering \cite{lukovnikov2017neural} and recommendation systems \cite{zhang2016collaborative}. A Knowledge Graph can be conceptualized as a directed graph, represented in the form of triples $(head\ entity, relation,\allowbreak  tail\ entity)$. 

\begin{figure}[t]
\centering
\includegraphics[width=0.6\textwidth]{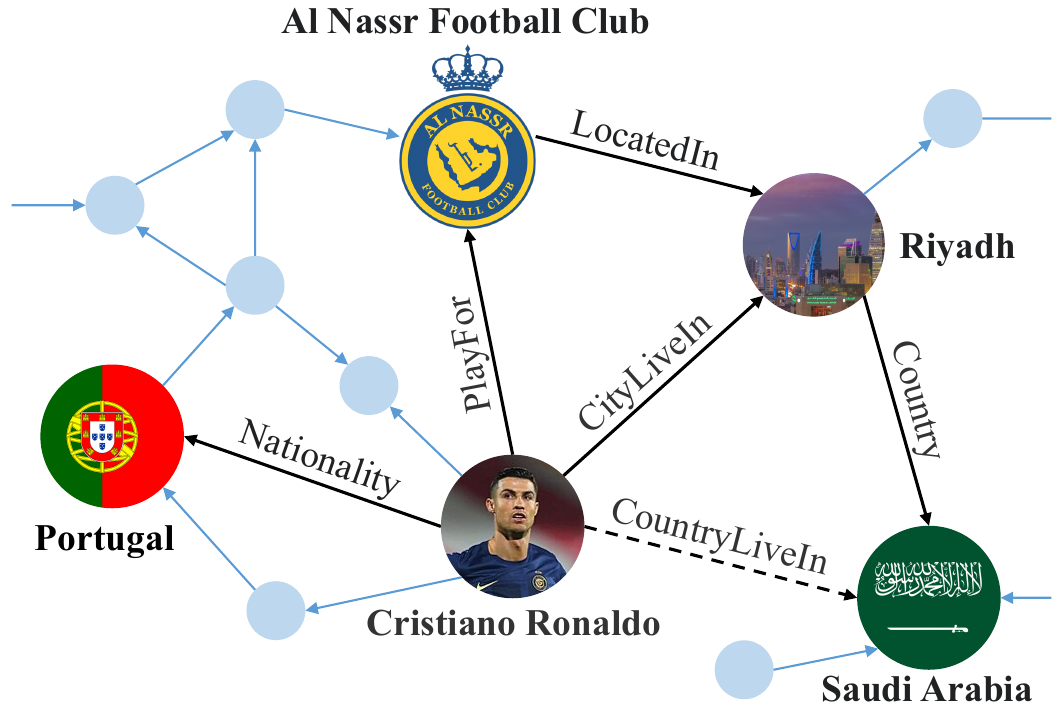}
\caption{Illustration of a Knowledge Graph with a potential link prediction task: inferring the missing triple (Cristiano Ronaldo, CountryLiveIn, Saudi Arabia) based on the existing facts}
\label{fig:dataset}
\end{figure}

However, KGs are commonly incomplete with many missing relations \cite{wang2014knowledge}. Link prediction, also known as Knowledge Graph Completion (KGC), is a form of automated reasoning used to infer missing parts of KGs. A popular approach for KGC is Knowledge Graph Embedding (KGE), which aims to project entities and relations into low-dimensional continuous vector space. Figure \ref{fig:dataset} provides a schematic illustration of a partial KG and how KGC infers potential missing relations between \textit{Cristiano Ronaldo} and \textit{Saudi Arabia}.

Current KGE approaches include translation-based models, semantic-matching models and neural network-based models. Translation-based models \cite{bordes2013translating, wang2014knowledge, sun2019rotate} establish linear translation rules between the head entity and the tail entity via the relation. Semantic-matching \cite{nickel2011three, yang2015embedding, trouillon2016complex, balavzevic2019tucker} models employ various score functions to measure the embedding similarity between entities and relations. However, they are facing an issue that the high-dimensional embeddings required to encode all the information when dealing with large KG, which can lead to overfitting and increased complexity. And they usually learn a single static representation. However, entities and relations should have different meanings when involved in different contexts. Besides, they mainly rely on additive or multiplicative operations. These factors limit their expressiveness \cite{chen2020knowledge}.

Neural network-based models \cite{dettmers2018convolutional, wang2019coke} obtain expressive representations from pure embeddings using different neural network models which have achieved good results on KGC task. One of them is Transformer-based model. Transformer \cite{vaswani2017attention} can learn the representation of entities and relations and contextual information of KG, thus it is suitable for link prediction tasks. CoKE \cite{wang2019coke} is one of the representative models. Although Transformer-based models have outperformed previous models with similarly high-dimensional embeddings, they still face scalability issues. 

To address these limitations, we propose a novel architecture called Triple Receptance Perception (TRP), which is specifically designed to capture dynamic and sequential contexts in knowledge graphs. Unlike conventional static methods, TRP dynamically models the context of entities and relations by incorporating sequential information directly into the representation learning process. This mechanism ensures that the embeddings are adaptively conditioned on varying graph contexts, thereby enhancing the model's capacity to encode complex relational dependencies.

In addition, to ensure efficient relational decoding, we integrate TRP with a Tucker decomposition module for triple scoring. Tucker decomposition, used widely in machine learning, can decompose a tensor into a set of matrices and a smaller core tensor. In knowledge graph, entity and relation embeddings can be modeled as components of the decomposed tensor, allowing for efficient representation learning and capturing interactions in a structured manner. This combination allows for a compact yet expressive representation of the relational structure in knowledge graphs, balancing parameter efficiency with performance.

We validate our method through experiments on benchmark datasets, including YAGO3-10, UMLS, FB15k, and FB13, where it consistently outperforms several state-of-the-art models in both link prediction and triple classification tasks. 

The paper is organized as follows: Section 2 outlines the related work. Section 3 details the proposed method, including the TRP encoder and Tucker decomposition decoder. The experimental results are presented in Section 4. Finally, Section 5 provides the conclusion.

\section{Related Work}\label{sec2}
One mainstream approach for Knowledge Graph Completion (KGC) is based on Knowledge Graph Embedding (KGE) methods \cite{chen2020knowledge}. The goal of KGE is to embed the representations of entities and relations into low-dimensional continuous vector space. KGE-based methods can be broadly classified into the following three types: translation-based models, semantic-matching models and neural networks-based models.

Translation-based models establish linear translation rules between the head entity and the tail entity via the relation. TransE \cite{bordes2013translating}, the most representative translation-based model, represents entities and relationships in the same space as vectors. The core idea of TransE is the sum of the head entity vector and the relationship vector should be as close as possible to the tail entity vector, thereby completing the representation of all entities and relationships. Although TransE is simple and efficient, it involves issues when modeling complex relations. Thus, TransH \cite{wang2014knowledge}, TransR \cite{lin2015learning}, TransD \cite{ji2015knowledge} and TranSparse further handle the issue when facing 1-N, N-1 and N-N relations by transforming entities and relations into different subspaces. RotateE \cite{sun2019rotate} defining each relations as a rotation from the head entity to the tail entity in the complex vector space. HAKE \cite{zhang2020learning}, a model that embeds the entities into a polar coordinate system, aims to capture the semantic hierarchy in KGs, thus making entities at different levels of the hierarchy distinguishable. HousE \cite{li2022house} utilizes the Householder parameterisation to capture crucial relation patterns.

Semantic-matching models employ various score functions to measure the embedding similarity between entities and relations. RESCAL \cite{nickel2011three} is a representative tensor decomposition model. It captures pairwise interactions between entities and relations based on a three-way tensor factorization. DistMult \cite{yang2015embedding} simplifies the RESCAL by changing the relation matrix to a diagonal matrix, reducing the number of training parameters, but it can only handle symmetric relations. ComplEx \cite{trouillon2016complex} extends matrix decomposition to the complex space, thus better handling asymmetric relations. ANALOGY introduces linear mapping to capture the analogical properties between entities and relations. TuckER \cite{balavzevic2019tucker} adopts Tucker decomposition to compute a validation score for each triple, and has gained promising results.

Neural networks-based models obtain expressive representations from pure embeddings using different neural network models. NTN \cite{socher2013reasoning} first proposed a simple feed-forward neural tensor network. ConvE \cite{dettmers2018convolutional}, ConvKB \cite{dai2018novel}, and InteractE \cite{vashishth2020interacte} use Convolutional Neural Networks (CNNs) to capture more feature interactions over embeddings for link prediction. Furthermore, Graph Neural Networks (GNNs) or Graph Attention Networks (GATs) models such as R-GCN \cite{schlichtkrull2018modeling} and CompGCN \cite{vashishth2019composition}, take advantage of rich information from structure and entities neighborhood to obtain better representations. Recently, Transformers have shown significant improvements in addressing various problems, including Knowledge Graph completion. KG-BERT \cite{yao2019kg} utilizes a pretrained BERT model to encode textual descriptions of Knowledge Graph triples, allowing it to infer new facts through a classification layer based on the special token representation. CoKE \cite{wang2019coke} employs a stack of Transformer encoders to encode the triples to obtain contextualized representations. HittER \cite{chen2021hitter} utilizes two levels of Transformer blocks, one to provide relation-dependent embeddings for an entity’s neighbors and the other to aggregate their information. 

In addition to the above KGE-based methods, other approaches have been proposed. Neural-LP \cite{yang2017differentiable} and DRUM \cite{sadeghian2019drum} propose a differentiable approach for rule-mining on KGs, which can learn logical rules and their related confidence scores. Large Language Models (LLMs) have shown great performance in natural language processing (NLP). Thus, some researchers attempt to unify LLMs and KGs together for KGC \cite{pan2024unifying}. ChatRule \cite{luo2023chatrule} employs LLMs to mine some critical logical rules. After ranking, these rules are used to conduct reasoning over KGs.

\section{Methodology}\label{sec3}
 
\begin{figure*}[t]
\centering
\includegraphics[width=0.95\textwidth]{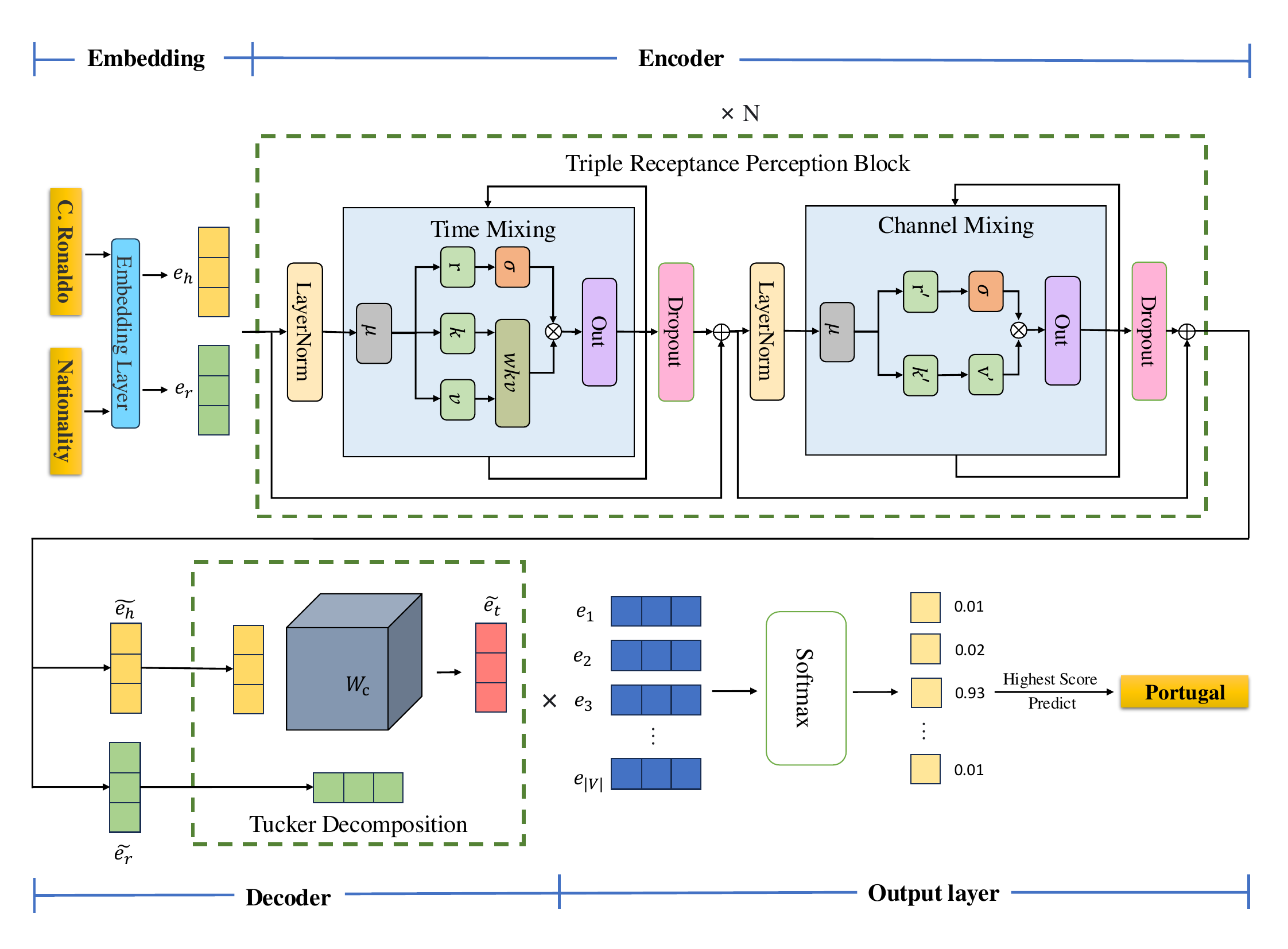}
\caption{Architecture of our method.}
\label{fig:architecture}
\end{figure*}

In this section, we present a novel approach to knowledge graph link prediction that combines two components: the Triple Receptance Perception (TRP) architecture for input encoding and Tucker decomposition for decoding and classification. The proposed methodology leverages TRP's capability to model sequential and contextual information dynamically, alongside robust Tucker decomposition capabilities for effective decoding. The model is trained to predict missing entities in knowledge graph triples by learning context-sensitive representations. Figure \ref{fig:architecture} provides a comprehensive overview of our methodology.

\subsection{Encoder}
\subsubsection{Triple Receptance Perception Architecture}
The Triple Receptance Perception (TRP) architecture is designed to address the challenges of capturing dynamic and sequential contexts in knowledge graphs. Unlike static embedding models that assign fixed representations to entities and relations, TRP allows for the learning of dynamic, context-dependent embeddings. This is achieved through its ability to model sequential information across knowledge graph triples, ensuring that the behavior of entities and relations is adapted to varying graph contexts. 
TRP consists of residual blocks, each primarily composed of two key sub-blocks: the time-mixing sub-block and the channel-mixing sub-block \cite{peng2023rwkv}. This architecture effectively combines the strengths of recurrent neural networks (RNNs) and attention mechanisms, enabling efficient integration of inductive or sequential information. 


\subsubsection{Time Mixing Module}
The operation of the time mixing module can be described as follows. Consider an input sequence $x = (e_{h}, e_{r})$, where $e_{h}$ represents the embedding of head entities, $e_{r}$ represent the embedding of relations. The output embedding denoted as $o = (o_{h}, o_{r})$, captures the contextual information and dependencies within the input sequence. The output is computed using the following equation, 
\begin{equation}
o_t = W_o \cdot (\sigma(r_t) \odot wkv_t),
\end{equation}
where $t \in \{h, r\}$ and $W_o$ is a weighting matrix for output vectors. Note that $\sigma(r_t)$ is the sigmoid of \textit{receptance} vector, and $r_t$ is calculated as:
\begin{equation}
r_t = W_r \cdot (\mu_r \odot x_t + (1 - \mu_r)\odot x_{t-1})
\end{equation}
In this equation $x_t$ stands for input embedding at step t and $\mu_r$ is a interpolation factor. The term $wkv_t$ aligns with the methods used in the Attention-Free Transformer\cite{zhai2021attention}, and can be expressed as

\begin{equation}
wkv_t = \frac{\sum_{i=1}^{t-1} e^{-(t-1-i)w+k_i} \odot v_i + e^{u+k_t} \odot v_t}{\sum_{i=1}^{t-1} e^{-(t-1-i)w+k_i} + e^{u+k_t}},  \label{eq:wkv}
\end{equation}
where $w$ is the channel-wise learnable vector for the previous input, while $u$ is the specialized weighting factor applied to the current input. $k$, $v$ are from \textit{K}, \textit{V} respectively. The primary role of $u$ is to provide a distinct attention channel for the current step, thereby mitigating the potential degradation issues associated with $w$. Consequently, the model learns the vector $u$ to better balance and capture information across different positions.
The \textit{key} and \textit{value} vectors are calculated as
\begin{equation}
k_t = W_k \cdot (\mu_k \odot x_t + (1 - \mu_k) \odot x_{t-1}),
\end{equation}
\begin{equation}
v_t = W_v \cdot (\mu_v \odot x_t + (1 - \mu_v) \odot x_{t-1}).
\end{equation}
Similarly, $\mu_k$ and $\mu_v$ are interpolation factors, while $W_k$ and $W_v$ are the weighting matrices for the key and value, respectively. Regarding the dimensions of these matrices, let $d_{io}$ denote the input/output size and $d_{att}$ represent the size of time-mixing module. The projection matrix $W_o \in \mathbb{R}^{d_{io} \times d_{att}}$ is defined for output projections. The projection matrices for receptance, key, and value are $W_r \in \mathbb{R}^{d_{att} \times d_{io}}$, $W_k \in \mathbb{R}^{d_{att} \times d_{io}}$, and $W_v \in \mathbb{R}^{d_{att} \times d_{io}}$, respectively.

Note that $wkv_t$ is the weighted summation of the input in the interval $[1, t]$, which permits the causality in inference and enables efficient inference like RNNs. Additionally, Eq.\eqref{eq:wkv} can be calculated recursively,
\begin{align}
    wkv_t &= \frac{a_{t-1} + e^{u+k_t} \odot v_t}{b_{t-1} + e^{u+k_t}} \\
    a_t &= e^{-w} \odot a_{t-1} + e^{k_t} \odot v_t \\
    b_t &= e^{-w} \odot b_{t-1} + e^{k_t} \\
    a_0, b_0 &= 0.
\end{align}

\subsubsection{Channel Mixing Module}
In the channel mixing module, the modified input sequence $x' = (x'_1, x'_2, \dots, x'_T)$ undergoes transformation as
\begin{align}
    r'_t &= W_{r'} \cdot (\mu'_r x'_t + (1 - \mu'_r) x'_{t-1}), \\
    k'_t &= W_{k'} \cdot (\mu'_k x'_t + (1 - \mu'_k) x'_{t-1}), \\
    o'_t &= \sigma(r'_t) \cdot (W_{v'} \odot \max(k'_t, 0)^2),
\end{align}
where $\mu'_r, \mu'_k$ are interpolation factors and $W_{r'}, W_{k'}, W_{r'}$ are separate weighting matrix for transformed vectors. We adopt squared ReLU activation function here for the output.

\subsubsection{Integration of Modules}
Each block processes the input by sequentially applying a dropout-enhanced time mixing followed by a channel mixing operation, formulated as
\begin{equation}
x' = x + \text{Dropout}(\text{TimeMixing}(\text{LayerNorm}(x))),
\end{equation}
\begin{equation}
x'' = x' + \text{Dropout}(\text{ChannelMixing}(\text{LayerNorm}(x'))). 
\end{equation}
This design introduces dropout layers prior to residual connections to mitigate overfitting issues.

These components collectively enable TRP architecture to handle sequences effectively, maintaining causality and enhancing model robustness through its recurrent-inspired mechanism.

\subsection{Decoder}

Tucker decomposition is a method that decomposes a high-dimensional tensor into the product of a low-dimensional core tensor and multiple matrix factors. In our model, we employ Tucker decomposition as the decoder.

In the knowledge graph completion task, the knowledge graph is typically represented as a triplet set $(e_h, r, e_t)$, where $e_h$ is the head entity, $r$ is the relation, and $e_t$ is the tail entity.  After obtaining the output representations of the head entity and the relation, we can construct a third-order tensor, where each element is a triplet tuple with a value of 1 if the triplet is true, and 0 otherwise. We then apply Tucker decomposition, a tensor factorization method, to decode the embedded information. Let $H_o = (\tilde{e}_h; \tilde{e}_r)$ denote the encoder output representation, where $\tilde{e}_h \in \mathbb{R}^d$ and $\tilde{e}_r \in \mathbb{R}^d$ are the output representations of the head entity and relation, respectively. The scoring function in decoder is defined as:
\begin{equation}
\phi(h, r, t) = W_c \times_1 \tilde{e}_h \times_2 \tilde{e}_r \times_3 e_t, 
\end{equation}
where $W_c \in \mathbb{R}^{d \times d \times d}$ is the core tensor of Tucker decomposition, which can be learned, and $\times_n$ denotes the n-mode tensor product.

\subsection{Training}
During training, we implement reciprocal learning by adding inverse triples (\(t, r^{-1}, h\)) to our dataset. This approach enhances the model's understanding of relationships in both directions. Additionally, we employ the 1-N scoring method, where each pair \((h, r)\) is evaluated against all possible entities as the target \(t\). This scoring mechanism is crucial for effectively handling the large scale of entity combinations.

To measure the predictive accuracy and improve the model's performance, we employ the cross-entropy loss function between the label and the prediction as our training loss:
\begin{equation}
\mathcal{L}(x) = -\sum_t y_t \log p_t,
\end{equation}
where \(y_t\) denotes the true label's probability distribution, one-hot encoded with a '1' for the correct class and '0s' elsewhere. The predicted probabilities \( p_t \) are computed using the softmax function applied to the logits \( \phi(h, r, t) \) output by the model. The softmax function is defined as follows:
\begin{equation}
p_t = \frac{e^{\phi(h, r, t)}}{\sum_{k \in |V|} e^{\phi(h, r, k)}},
\end{equation}
where \(|V|\) denotes the number of all entities.

\section{Experiments}

\subsection{Experiment setup} 
\subsubsection{Datasets} 
We evaluate our model on link prediction task using both small and large datasets (dataset statistics are provided in Table~\ref{tab:dataset_statistics}):

\textbf{UMLS} \cite{mccray2003upper} is the Unified Medical Language System (UMLS), developed by McCray, and is a comprehensive resource from the biomedical domain. The entities are concepts of drug and disease names and the relations are concepts of diagnosis and treatment. It serves as a small knowledge graph. 

\textbf{FB15k} \cite{bordes2013translating} is a subset of Freebase, a large database containing real-world facts. The entities encompass concepts such as movies, actors, awards, and sports. It has the largest number of relations, that is 1,345.

\textbf{YAGO3-10} \cite{mahdisoltani2013yago3} is a subset of YAGO3, including entities associated with 46 relations. Most of the triples describe attributes of persons, such as citizenship, gender, and profession. This dataset is significantly larger than the others.

\textbf{FB13} \cite{socher2013reasoning} is a subset of Freebase, including 75,043 entities and 13 relations. The validation and test datasets consist of negative triples, which are used for the triple classification task.

\begin{table*}[t]
\centering
\caption{Dataset Statistics}
\label{tab:dataset_statistics}
\begin{tabular}{l c c ccc}
\toprule
\textbf{Dataset} & \textbf{\#Entity} & \textbf{\#Relation} & \multicolumn{3}{c}{\textbf{\#Triplet}} \\
\cmidrule(lr){4-6}
& & & Train & Valid & Test \\
\midrule
UMLS & 135 & 46 & 5,126 & 652 & 661 \\
FB15k & 14,951 & 1,345 & 483,142 & 50,000 & 59,071 \\
YAGO3-10 & 123,182 & 37 & 1,079,040 & 5,000 & 5,000 \\
FB13 & 75,043 & 13 & 316,232 & 11,816 & 47,466 \\
\bottomrule
\end{tabular}%
\end{table*}

\subsubsection{Baseline}
For fairness, we do not compare with models that utilize auxiliary information like text description. We choose several representative methods, which can be divided into two categories: methods using triples and methods using context. 

Methods using triples rely exclusively on the structural information present in the knowledge graph, focusing on the relationships between entities as defined by the triples. For comparison, we include TransE \cite{bordes2013translating}, DistMult \cite{yang2015embedding}, ComplEx \cite{trouillon2016complex}, RotatE \cite{sun2019rotate}, TuckER \cite{balavzevic2019tucker}, ConvE \cite{dettmers2018convolutional}, CoKE \cite{wang2019coke}, HAKE \cite{zhang2020learning}, and HousE \cite{li2022house} as our baselines.

In contrast, methods using context leverage additional information, such as graph structure or logic rules, during the training process.
For these, we include Neural-LP \cite{yang2017differentiable}, R-GCN \cite{schlichtkrull2018modeling}, Rlogic \cite{cheng2022rlogic}, and ChatRule \cite{luo2023chatrule} as our baselines.

\subsubsection{Evaluation Metrics} 
For each triplet $(h, r, t) \in S$, where $S$ is the validation or testing set, we compute the score of $(t, r^{-1}, h')$ for all $h' \in E$ and determine the rank of $h$. Similarly, we compute the scores of $(h, r, t')$ for all $t' \in E$ to determine the rank of $t$. The relation $r$ is not compared.

Following previous works, we evaluate our model for link prediction using two standard metrics:

\textbf{Mean Reciprocal Rank (MRR)}: This metric is the average of the reciprocal ranks of the correct triples among all candidate triples:
\begin{equation}
  \text{MRR} = \frac{1}{|S|} \sum_{i=1}^{|S|} \frac{1}{\text{rank}_i},
\end{equation}
  where $\text{rank}_i$ is the rank assigned to the $i$-th true triple.

\textbf{Hits@k}: This metric measures the proportion of times the true triple is ranked within the top $k$ candidate triples:
\begin{equation}
  \text{Hits@k} = \frac{1}{|S|} \sum_{i=1}^{|S|} \mathbb{I}(\text{rank}_i \leq k),
\end{equation}
  where $\mathbb{I}(\cdot)$ is the indicator function, which returns 1 if the condition is true and 0 otherwise.

In the filtered setting \cite{bordes2013translating}, we exclude any candidate triples that are already present in the training set when evaluating the rank of the true triple.

\subsubsection{Training Details}
We implement our model based on the PyTorch library and conduct all experiments with a single NVIDIA RTX 4090 GPU
Our settings for hyper-parameter selection as follows: 
The embedding size \( k \) is selected in \(\{64, 96, 128, 192, 256\}\).
The number of TRP blocks is selected in \(\{2, 4, 6, 8\}\).
We employ dropout on all layers, with the rate tuned in \( \rho \in \{.2, .3, .4, .5\} \).
We use the Adam optimizer.
The learning rate is chosen from 0.0005 to 0.01, and different learning rates can be selected according to different datasets.
We train with batch size \( B = 512 \) for at most 500 epochs. The best hyper-parameter setting on each dataset is determined by MRR on the dev set.

\subsection{Results}

\subsubsection{Link Prediction Results}

\begin{table*}[t]
\centering
\caption{Performance comparison of various methods for link prediction on FB15k, YAGO3-10, and UMLS datasets. Best results are highlighted in \textbf{bold}, and second-best results are \underline{underlined}. `-' means that we failed to access experimental results from their original code or paper. }
\label{tab:link_prediction_result}
\resizebox{0.98\textwidth}{!}{
\begin{tabular}{l l ccc ccc ccc}
\toprule
\multirow{2}{*}{}& \multirow{2}{*}{Method} & \multicolumn{3}{c}{FB15k} & \multicolumn{3}{c}{YAGO3-10} & \multicolumn{3}{c}{UMLS} \\
\cmidrule(lr){3-5} \cmidrule(lr){6-8} \cmidrule(lr){9-11}
& & mrr & h@1 & h@10 & mrr & h@1 & h@10 & mrr & h@1 & h@10 \\
\midrule
\multirow{6}{*}{\makecell[c]{Methods \\ using triples}} 
& TransE & 0.38 & 23.1 & 47.1 & 0.30 & 21.8 & 47.5 & 0.69 & 52.3 & 89.7 \\
& DistMult & 0.65 & 54.6 & 82.4 & 0.37 & 26.2 & 57.5 & \underline{0.94} & \textbf{91.6} & 99.2 \\
& ComplEx & 0.69 & 59.9 & 84.0 & 0.42 & 32.0 & 60.3 & 0.92 & 85.5 & \underline{99.7} \\
& RotatE & 0.79 & 74.6 & 88.4 & 0.50 & 40.2 & 67.0 & - & - & - \\
& TuckER & 0.79 & 74.1 & 89.2 & 0.47 & 38.0 & 64.7 & - & - & - \\
& ConvE & 0.75 & 67.0 & 87.3 & 0.52 & 45.0 & 66.0 & 0.92 & 88.0 & 99.2 \\
& CoKE & \textbf{0.85} & \textbf{82.6} & \textbf{90.6} & \underline{0.55} & 47.5 & 67.5 & \underline{0.94} & \underline{90.7} & \underline{99.7} \\
& HAKE & - & - & - & \underline{0.55} & 46.2 & 69.4 & - & - & - \\
& HousE & \underline{0.81} & 75.9 & 89.8 & \textbf{0.57} & \underline{49.1} & \textbf{71.4} & - & - & - \\
\midrule
\multirow{4}{*}{\makecell[c]{Methods \\ using context}} 
& Neural-LP & 0.76 & - & 83.7 & - & - & - & 0.72& 58.2& 93.0\\
& R-GCN & 0.70 & 60.1 & 84.2 & 0.24 & 21.2 & 42.1 & - & - & - \\
& Rlogic & 0.31 & 20.3 & 50.1 & 0.36 & 25.2 & 50.4 & 0.71 & 56.6 & 93.2 \\
& ChatRule & - & - & - & 0.45 & 35.4 & 62.7 & 0.78 & 68.5 & 94.8 \\
\midrule
& ours & \textbf{0.85} & \underline{81.2} & \underline{90.3} & \textbf{0.57} & \textbf{50.1} & \underline{70.0} & \textbf{0.95} & 90.4 & \textbf{99.9} \\
\bottomrule
\end{tabular}
}
\end{table*}

Link prediction aims to predict missing head or tail entity of a triple, which is a widely employed evaluation task for knowledge graph completion models. Table ~\ref{tab:link_prediction_result} showcases the performance of our proposed method relative to existing representative models on FB15k, YAGO3-10, and UMLS datasets. Our method stands out for its performance, consistently outperforming both methods using triples and methods using context.
On FB15k, although our method slightly lags behind the transformer-based CoKE in Hit@1 and Hit@10 metrics, the gap is minimal, and our model consistently outperforms the other competing models across the board.
On the large YAGO3-10 dataset, our approach achieved the best results, demonstrating its effectiveness on datasets with a high volume of entities. This success underscores the robustness of our method in handling the complexity and scale of large-scale knowledge graphs.
On the smaller UMLS dataset, our approach also delivered outstanding performance, achieving the highest scores across all metrics. This indicates that our method is not only effective for large-scale datasets but also excels with smaller datasets, showcasing its ability to handle diverse data sizes efficiently.

\subsubsection{Triple Classification}

\begin{table*}[t]
\centering
\caption{Performance comparison of various methods for triple classification on the FB13 and FB15k datasets. Best results are highlighted in \textbf{bold}, and second-best results are \underline{underlined}.}
\label{tab:triple_classification_result}

\begin{tabular}{l c c}
\toprule
\textbf{Method} & \textbf{FB13} & \textbf{FB15K} \\
\midrule
NTN & 87.1 & 84.5 \\
TransE & 81.5 & 79.8 \\
TransH & 83.3 & 80.2 \\
TransR & 82.5 & 83.9 \\
DistMult & 86.2 & 85.1 \\
ComplEx & 85.7 & 86.2 \\
CoKE & \underline{87.7} & \textbf{89.3} \\
\midrule
ours & \textbf{88.6} & \underline{89.0}\\
\bottomrule
\end{tabular}

\end{table*}

Triple classification aims to judge whether a given triple $(h, r, t)$ is correct or not. This is a binary classification task, which has been explored in \cite{socher2013reasoning, wang2014knowledge} for evaluation. In this task, we use two datasets, FB13 and FB15k.
We need negative triples for evaluation of binary classification. The datasets FB13 released by NTN \cite{socher2013reasoning} already have negative triples, which are obtained by corrupting correct triples. As FB15k with negative triples has not been released by previous work, we construct negative triples following the same setting in \cite{socher2013reasoning}. For triple classification, we set a relation-specific threshold $\delta_r$. For a triple $(h, r, t)$, if the dissimilarity score obtained by $f_r$ is below $\delta_r$, the triple will be classified as a false fact, otherwise it will be classified as a true fact. And different values of $\delta_r$ will be set for different relations. We use the same settings as link prediction task, all parameters are optimized on the validation datasets to obtain the best accuracies. We compare our method NTN  \cite{socher2013reasoning}, TransE \cite{bordes2013translating}, TransH \cite{wang2014knowledge}, TransR \cite{lin2015learning}, Distmult \cite{yang2015embedding}, ComplEx \cite{trouillon2016complex}, CoKE\cite{wang2019coke}. The results are listed in Table \ref{tab:triple_classification_result}.
Experimental results show that our proposed model achieves the best performance on the FB13 dataset with an accuracy of 88.6\%, surpassing all other baseline methods, including Transformer-based method CoKE. On the FB15k dataset, our model also performs competitively, achieving an accuracy of 89.0\%, slightly below CoKE but outperforming other models. These results highlight the effectiveness of our model in capturing the underlying relationships between entities and relations in knowledge graphs. Hence, in the triple classification task, the proposed method can accurately judge whether the given triple is correct or not.

\subsubsection{Ablation Study}
We conducted an ablation study on FB15k and YAGO3-10 to evaluate the contributions of Tucker decomposition decoder and the TRP blocks. We performed experiments by removing the perception block encoder and Tucker decomposition decoder separately. When the Tucker decomposition decoder was removed, we used the embedding $\tilde{e}_r)$ output by TRP blocks directly as the final predicted entity embedding. When the TRP block encoder is removed, the model can be viewed as an original TuckER model \cite{balavzevic2019tucker}. Table ~\ref{tab:Ablation_result} shows the results of our ablation study on the FB15k and YAGO3-10. As shown, the combination of both modules yields the best results, confirming that both components are essential for achieving optimal performance.

\begin{table*}[t]
\centering
\caption{Ablation Study Results on FB15k and YAGO3-10. Best results are highlighted in \textbf{bold}.}
\label{tab:Ablation_result}
\begin{tabular}{l c c c c c c}
\toprule
\multirow{2}{*}{Ablation} & \multicolumn{3}{c}{\textbf{FB15k}} & \multicolumn{3}{c}{\textbf{YAGO3-10}} \\
\cmidrule(lr){2-4} \cmidrule(lr){5-7}
& MRR & H@1 & H@10 & MRR & H@1 & H@10 \\
\midrule
Original model & \textbf{0.85} & \textbf{81.2} & \textbf{90.3} & \textbf{0.57} & \textbf{50.1} & \textbf{70.0} \\
\midrule
\textit{w/o} Tucker Decomposition Decoder & 0.82 & 77.6 & 89.4 & 0.54 & 48.1 & 67.2 \\
\textit{w/o} TRP Encoder & 0.79 & 74.1 & 89.2 & 0.47 & 38.0 & 64.7 \\
\bottomrule
\end{tabular}
\end{table*}

\begin{table*}[t]
\centering
\caption{Comparison of Decoders with Score Functions on FB15k. Best results are highlighted in \textbf{bold}.}
\label{tab:Decoder_result}
\begin{tabular}{l c c c c}
\toprule
\multirow{2}{*}{\textbf{Decoders}} & \multirow{2}{*}{\textbf{Score Function}} &\multicolumn{3}{c}{\textbf{FB15k}} \\
\cmidrule(lr){3-5}
& & MRR & H@1 & H@10 \\
\midrule
Tucker decomposition
& $W_c \times_1 \mathbf{h} \times_2 \mathbf{r} \times_3 \mathbf{t}$ & 
\textbf{0.85} & \textbf{81.2} & \textbf{90.3} \\

\midrule
MLP & $\langle \mathbf{W}_2 \text{ReLU}(\mathbf{W}_1 [\mathbf{h}, \mathbf{r}] + \mathbf{b}_1) + \mathbf{b}_2, \mathbf{t} \rangle$ & 0.83 & 80.0 & 89.5 \\
TransE \cite{bordes2013translating} & $-\left\|\mathbf{h} + \mathbf{r} - \mathbf{t}\right\|$ & 0.79 & 73.3 & 88.2 \\
DistMult \cite{yang2015embedding} & $\langle \mathbf{r}, \mathbf{h}, \mathbf{t} \rangle$ & 0.79 & 73.9 & 88.4 \\
ComplEx \cite{trouillon2016complex} & $\text{Re} (\langle \mathbf{r}, \mathbf{h}, \mathbf{t} \rangle)$ & 0.83 & 78.0 & 90.1 \\
\bottomrule
\end{tabular}
\end{table*}

And we also conduct an ablation study to evaluate the impact of different decoders on the performance of knowledge graph completion. We perform experiments on FB15k with the following decoders: MLP, TransE, DistMult and ComplEx. To ensure a fair comparison, we keep all other experimental settings as consistent as possible. As shown in Table \ref{tab:Decoder_result}, the results shows that our model with Tucker decomposition decoder achieves the best performance. This demonstrates the effectiveness of Tucker decomposition as decoder, which is able to capture more complex interactions between entities and relations.

\begin{figure}[t]
  \centering
  \begin{subfigure}[b]{0.48\textwidth}
    \includegraphics[width=\textwidth]{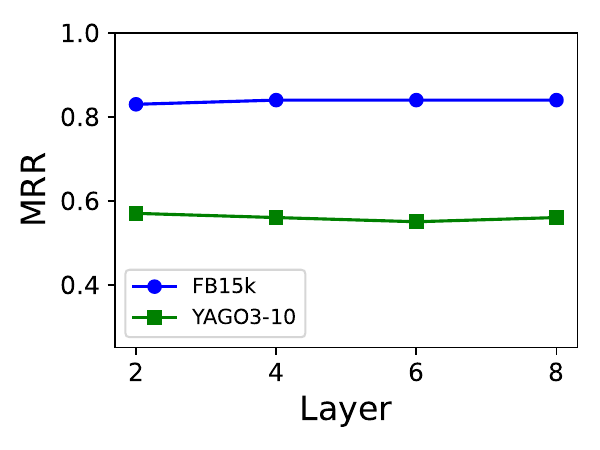}
    \label{fig:layer_result}
  \end{subfigure}
  \hfill 
  \begin{subfigure}[b]{0.48\textwidth}
    \includegraphics[width=\textwidth]{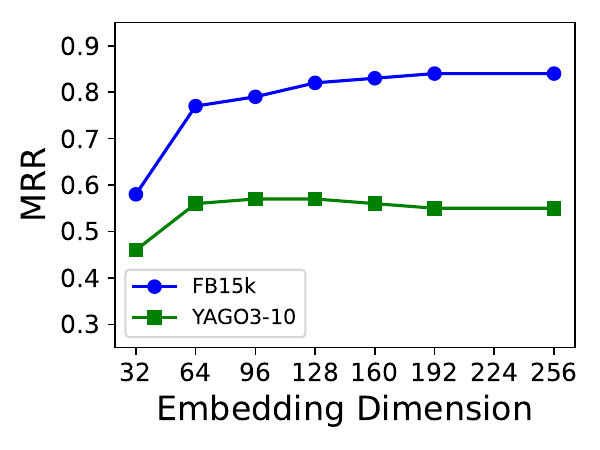}
    \label{fig:emb_result}
  \end{subfigure}
  \caption{Effect of the number of layers (Left) and embedding dimension (Right) on MRR for FB15k and YAGO3-10}
  \label{fig:layer_emb_result}
\end{figure}

\subsubsection{Parameter Sensitivity Analysis}
In this section, we aim to explore the impact of different hyperparameters on the overall performance of the model. Specifically, we investigated how the number of TRP block layers in perception block encoder and the dimensions of the embedding affect the model's performance.

We begin by analyzing the effect of varying the number of TRP block layers. We experimented with models containing 2, 4, 6, 8, 10 layers, respectively. Each model was trained separately, and MRR and Hits@10 were used as the evaluation metrics. We observed that the number of TRP block layers had minimal impact on model performance, as shown in the Figure \ref{fig:layer_emb_result}. The MRR remained relatively stable across different layer configurations, indicating that increasing the number of TRP block layers does not significantly improve the model's performance in the current experimental setup. Therefore, using fewer layers, such as 2 or 4, may be sufficient to achieve good performance while keeping computational costs low.

Next, we analyze the impact of embedding dimension size on the model's performance. Embedding dimensions, which map features of entities and relations into a lower-dimensional space, are crucial hyperparameters. Variations in dimension size can significantly affect model performance. We conducted experiments across various embedding dimensions, as shown in the Figure \ref{fig:layer_emb_result}. On the FB15k dataset, the model reached near-optimal performance at 128 dimensions, while on YAGO3-10, the model achieved its best performance starting from 64 dimensions. This demonstrates that the model is able to work well with relatively small embedding sizes.

\begin{figure}[ht]
  \centering
  \begin{subfigure}[b]{0.48\textwidth}
    \includegraphics[width=\textwidth]{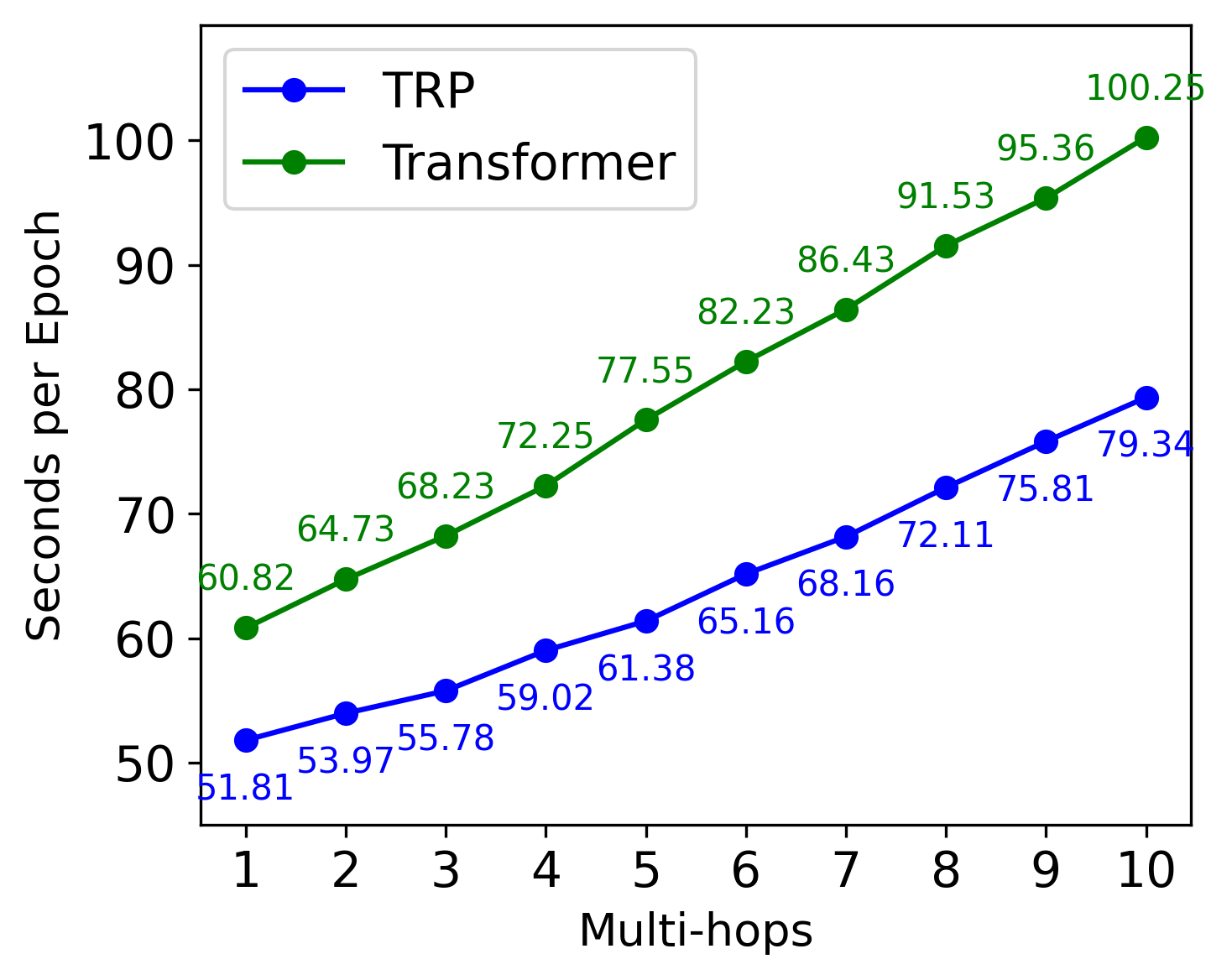}
    \label{fig:runtime}
  \end{subfigure}
  \hfill 
  \begin{subfigure}[b]{0.48\textwidth}
    \includegraphics[width=\textwidth]{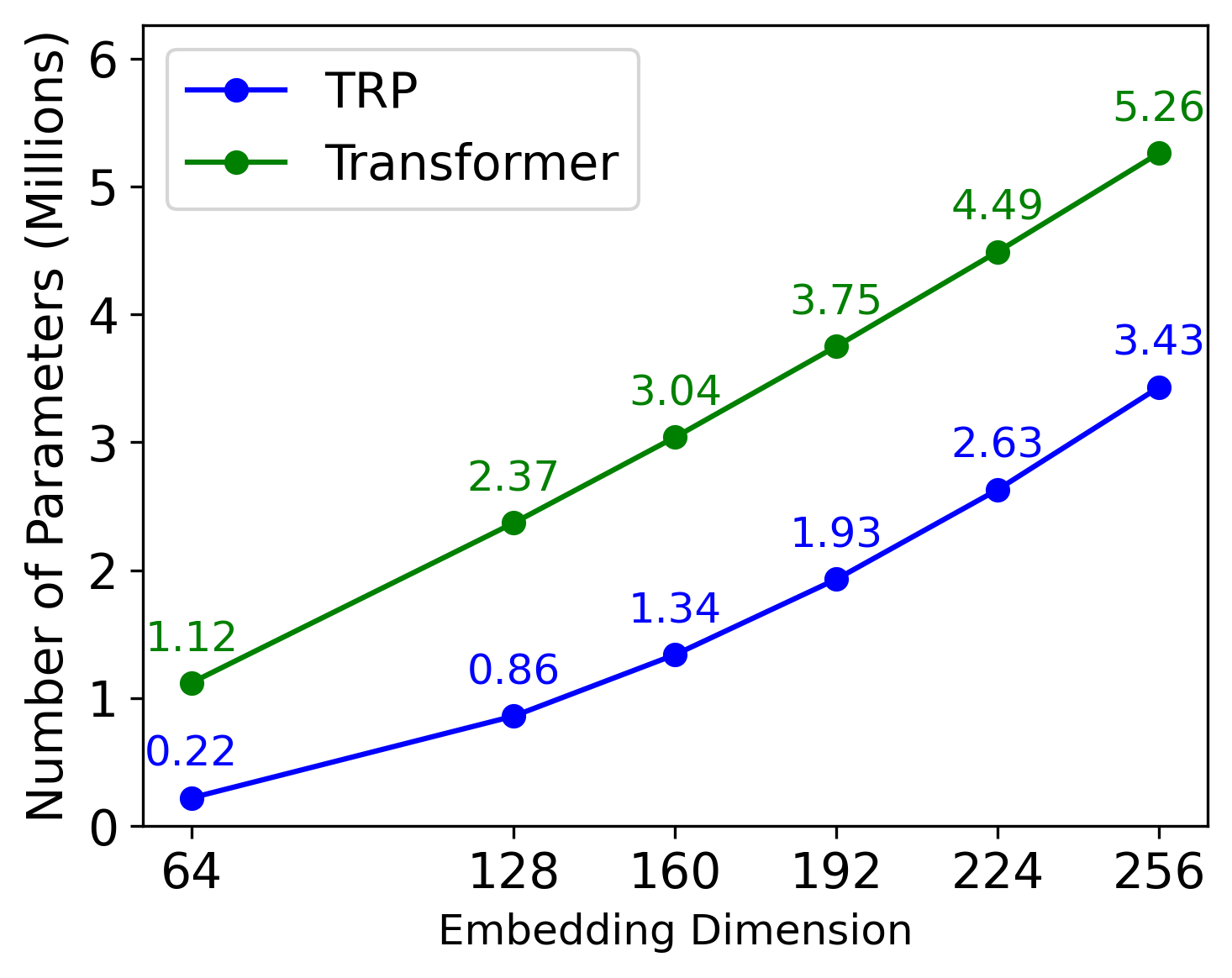}
    \label{fig:parameters}
  \end{subfigure}
  \caption{Training time per epoch across dataset proportion on YAGO3-10 (Left) and parameter efficiency comparison between TRP and Transformer across different embedding dimensions. (Right)}
\end{figure}

\subsubsection{Parameter Efficiency}
In this section, we investigate the parameter efficiency of the TRP architecture. To evaluate this, we compare TRP with Transformer, one of the most effective sequence models. For a fair comparison, both models are configured with the same number of layers.

In our experiments, we simulate the performance of TRP and Transformer under multi-hop scenarios. Specifically, the input for the multi-hop setting follows the structure: $\textit{head\ entity} - \textit{relation}_1 - \textit{entity}_2 - \textit{relation}_2 - \dots - \textit{entity}_n - \textit{relation}_n$. For the training dataset, we utilize YAGO3-10 and ensure the number of samples in the multi-hop training dataset remains consistent across experiments. We evaluate their performance under different hop counts, and the results are presented in Figure~\ref{fig:runtime}. As shown, TRP requires less training time per epoch compared to Transformer, and the growth in training time is also slower as the number of hops increases.

Additionally, we analyze the parameter counts across different embedding dimensions on the YAGO3-10 dataset, as illustrated in Figure~\ref{fig:parameters}. The results clearly demonstrate that models employing TRP encoder require significantly fewer parameters than Transformer. Furthermore, based on our experimental observations, both models achieve comparable performance in link prediction tasks, with TRP slightly outperforming Transformer.

These experiments highlight the efficiency and effectiveness of the TRP encoder.

\subsubsection{Embedding Visualization}
\begin{figure}[t]
  \centering
  \begin{subfigure}[b]{0.43\textwidth}
    \includegraphics[width=\textwidth]{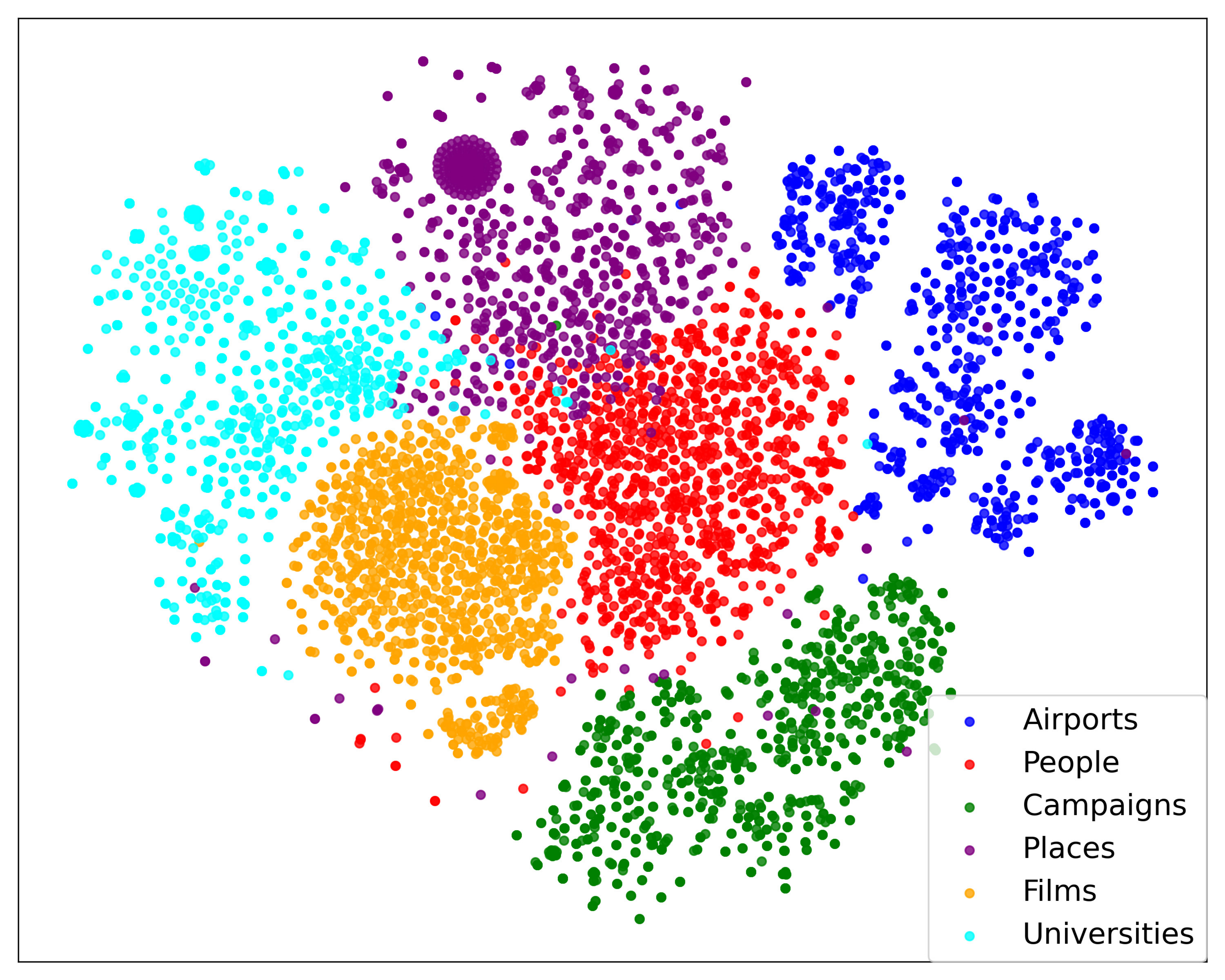}
    \caption{YAGO3-10 Embeddings}
    \label{fig:YAGO3-10_embedding}
  \end{subfigure}
  \hfill 
  \begin{subfigure}[b]{0.43\textwidth}
    \includegraphics[width=\textwidth]{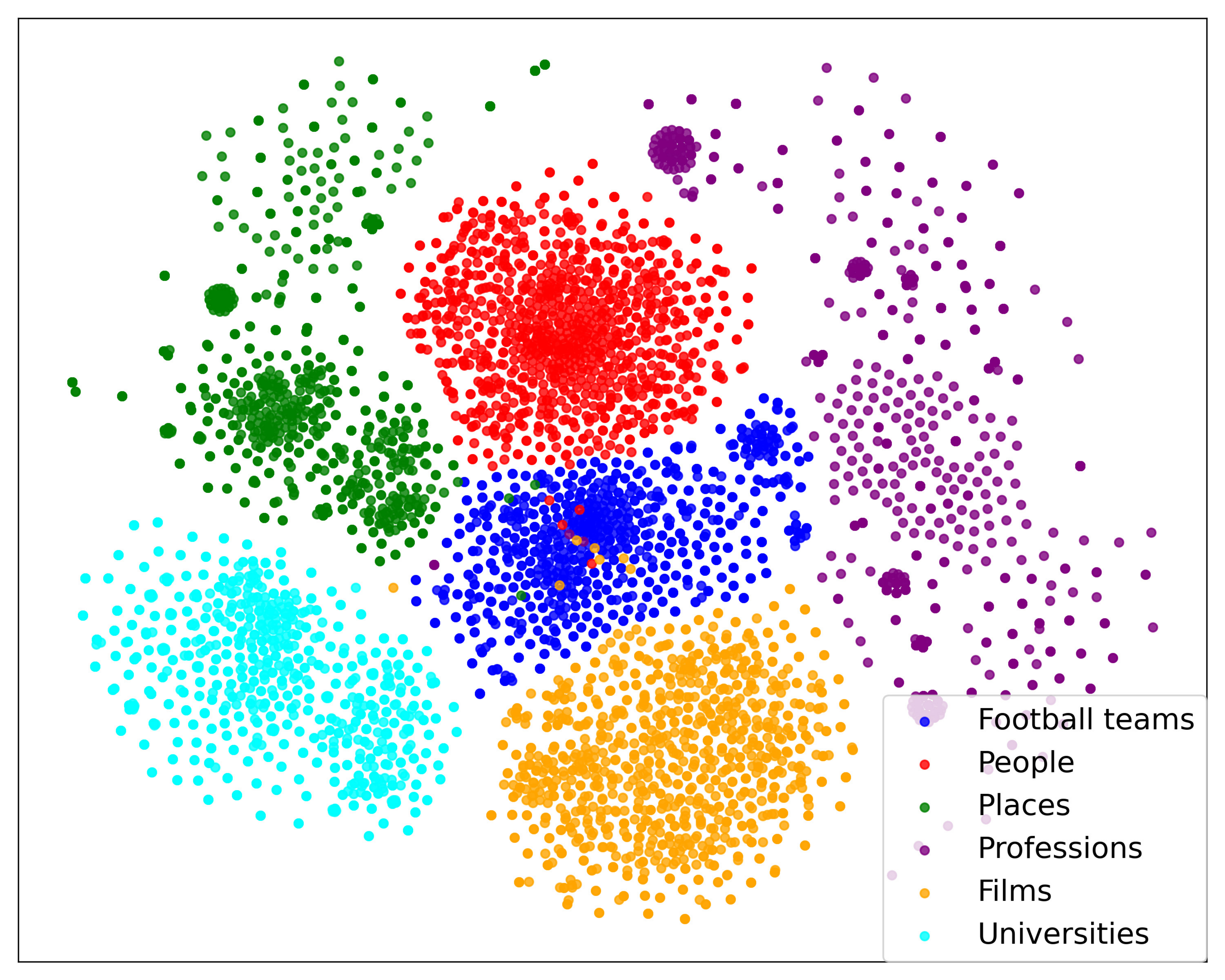}
    \caption{FB15k Embeddings}
    \label{fig:FB15k_embedding}
  \end{subfigure}
  \caption{t-SNE visualizations of entity embeddings from YAGO3-10 and FB15k. Each point represents an entity embedding, with different colors indicating various categories}
  \label{fig:embedding_visualizations}
\end{figure}

In this section, we visualize the embeddings generated by our knowledge graph completion model via t-SNE \cite{van2008visualizing}. We selected six entity categories from YAGO3-10 dataset—Airports, People, Campaigns, Places, Films, and Universities—and randomly sampled 1,000 entities from each category. We also selected six entity categories from FB15k dataset—Football teams, People, Places, Professions, Films and Universities—and randomly sampled 1,000 entities from each category. Each category is represented by a distinct color, helping to illustrate the distribution of different types of entities in the embedding space, as shown in Figure \ref{fig:embedding_visualizations}.
Both visualizations reveal that the model effectively separates the embeddings of most entity categories. Each category forms a distinct and compact cluster, clearly separated from other categories. Furthermore, within each category, the entities are tightly grouped together, indicating that our model has successfully captured the semantic relationships between entities within the same category in the knowledge graph.

\begin{figure}[t]
\centering
\includegraphics[width=.6\linewidth]{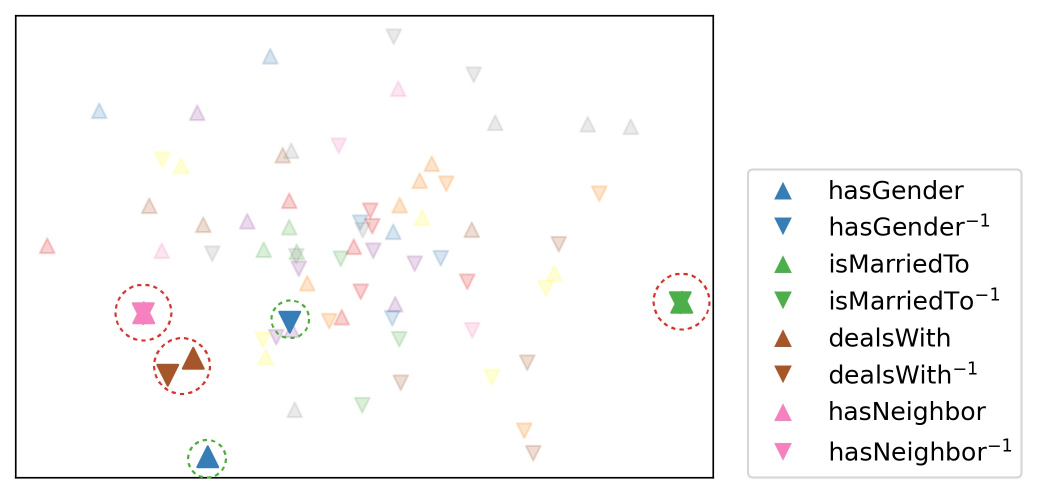}
\caption{t-SNE visualizations of relation embeddings \textbf{after TRP encoder} from YAGO3-10. Upward triangles denote positive relations, downward triangles indicate inverse relations, and relations in the same color correspond to a relation-inverse pair. We can observe that symmetric relations are closely clustered together. In contrast, asymmetric relations are distributed more distantly.}
\label{fig:relation_embedding}
\end{figure}

Additionally, we visualized the embeddings of different relations after TRP encoder from the YAGO3-10 dataset. For this visualization, we randomly selected a single entity and combined it with each relation, then passed them through TRP encoder to obtain the encoded relation embeddings. As shown in Figure \ref{fig:relation_embedding}, upward triangles denote positive relations, downward triangles indicate inverse relations, and relations in the same color correspond to a relation-inverse pair.
There are 37 kinds of relations in YAGO3-10 dataset. Notably, \textit{isMarriedTo}, \textit{hasNeighbor}, and \textit{dealsWith} are completely symmetric relations, meaning their inverse relations share identical semantics. In the figure, these relations are highlighted with red dashed circles. We observe that these relations and their inverse relation are closely clustered together. 
In contrast, for the relation \textit{hasGender}, shown in the green dashed circle in the figure, it is clear that this relation is not symmetric, as the embeddings are more dispersed. 
This indicates that different types of relations in the knowledge graph have distinct semantics, and TRP encoder has learned to embed them in separate regions to distinguish their individual roles. The wide dispersion also implies that TRP encoder is capable of capturing the diverse semantic meanings associated with various relations.

\section{Conclusion}
In this work, we propose a knowledge graph completion method that integrates the Triple Receptance Perception (TRP) architecture as the encoder and a Tucker decomposition module as the decoder. 
The TRP effectively models sequential information, enabling the learning of dynamic embeddings, while Tucker decomposition decoder provides robust relational decoding capabilities. 
We conduct experiments on link prediction and triple classification tasks, demonstrating that our method outperforms several state-of-the-art models. These results prove that the integration of TRP and Tucker decomposition decoder allows for more expressive representations. Our findings further confirm the efficiency and effectiveness of TRP. Visualizations further illustrate the effectiveness of our model in capturing the fine-grained contextual meanings of entities and relations.

\bibliography{sn-bibliography}

\end{document}